\def\year{2020}\relax
\documentclass[letterpaper]{article} % DO NOT CHANGE THIS
\usepackage{aaai20}  % DO NOT CHANGE THIS
\usepackage{times}  % DO NOT CHANGE THIS
\usepackage{helvet} % DO NOT CHANGE THIS
\usepackage{courier}  % DO NOT CHANGE THIS
\usepackage[hyphens]{url}  % DO NOT CHANGE THIS
\usepackage{graphicx} % DO NOT CHANGE THIS
\usepackage{graphicx}  % DO NOT CHANGE THIS
\usepackage{amsmath}
\usepackage{amssymb}
\usepackage{bm}
\usepackage{algorithm}  % added
\usepackage{algorithmicx} % added
\usepackage{algpseudocode} % added

\newcommand{\ie}{\textit{i}.\textit{e}.}
\newcommand{\eg}{\textit{e}.\textit{g}.}
\title{Image Cropping with Composition and Saliency Aware Aesthetic Score Map}
\author{Yi Tu,\textsuperscript{\rm 1}
Li Niu,\textsuperscript{\rm 1}\thanks{Corresponding author.}
Weijie Zhao,\textsuperscript{\rm 2} 
Dawei Cheng,\textsuperscript{\rm 1}
Liqing Zhang\textsuperscript{\rm 1}\footnotemark[1] \\
\textsuperscript{\rm 1}
MoE Key Lab of Artificial Intelligence, Department of Computer Science and Engineering\\
Shanghai Jiao Tong University, Shanghai, China\\
\{tuyi1991, ustcnewly, dawei.cheng\}@sjtu.edu.cn, zhang-lq@cs.sjtu.edu.cn\\
\textsuperscript{\rm 2}
Versa Inc, Shanghai, China\\
 weijie.zhao@versa-ai.com
}
 
\begin{document}

\maketitle

\begin{abstract}
Aesthetic image cropping is a practical but challenging task which aims at finding the best crops with the highest aesthetic quality in an image. Recently, many deep learning methods have been proposed to address this problem, but they did not reveal the intrinsic mechanism of aesthetic evaluation. In this paper, we propose an interpretable image cropping model to unveil the mystery. For each image, we use a fully convolutional network to produce an aesthetic score map, which is shared among all candidate crops during crop-level aesthetic evaluation. Then, we require the aesthetic score map to be both composition-aware and saliency-aware. In particular,  the same region is assigned with different aesthetic scores based on its relative positions in different crops. Moreover, a visually salient region is supposed to have more sensitive aesthetic scores so that our network can learn to place salient objects at more proper positions. Such an aesthetic score map can be used to localize aesthetically important regions in an image, which sheds light on the composition rules learned by our model. We show the competitive performance of our model in the image cropping task on several benchmark datasets, and also demonstrate its generality in real-world applications.
\end{abstract}
\section{Introduction}
\label{sec:introduction}

Given an image, the image cropping task aims at finding the crops with the best aesthetic quality. It is an important task that can be widely used in a lot of down-stream applications, \eg, photo post-processing \cite{chen2017learning}, view recommendation \cite{li2018a2,wei2018good}, and image thumbnailing \cite{esmaeili2017fast}. In order to find the best crop, an image cropping model will first generate a large number of candidate crops and then determine the best crop based on crop-level aesthetic evaluation. So an image cropping model is usually composed of two stages, candidate generation and aesthetic evaluation. 
A good image crop is achieved by selecting important contents and placing them with a good composition. The required knowledge for such a task can be categorized into two parts, \ie, content preference and composition preference. Therefore, a good image cropping model should be able to learn and leverage such preferences when searching for the best crop. 

\begin{figure}[tb]
    \centering
    \includegraphics[width=7cm]{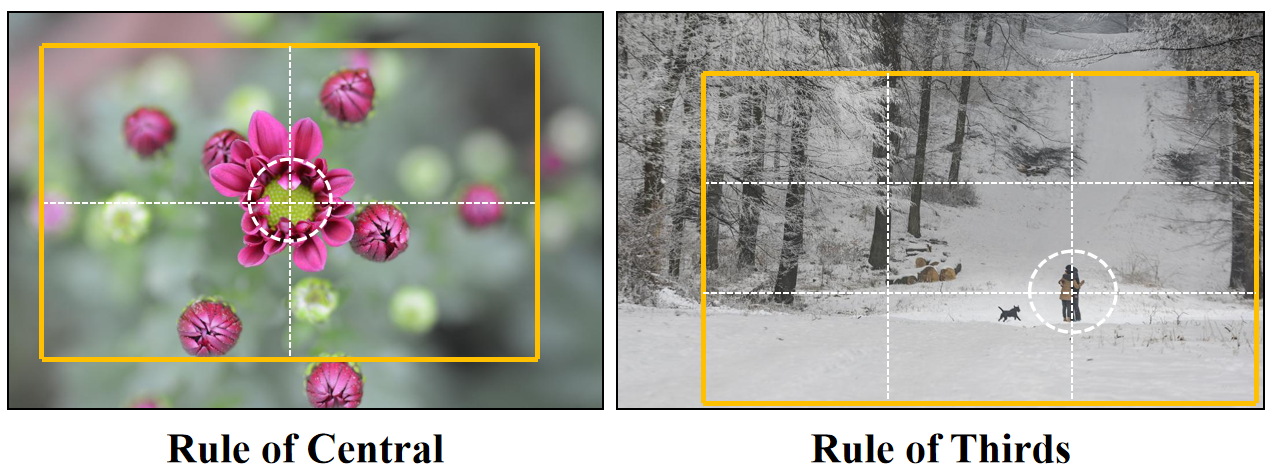}
    \caption{Images crop with composition rules.
The orange box in each image denotes a good crop found based on human-defined composition rules.
The white dotted lines denote the auxiliary lines used in these composition rules.}
    \label{fig:composition-rules}
\end{figure}

Early methods achieve this goal by explicitly utilizing some photography knowledge like human-defined composition rules, \eg, Rule of Thirds and Rule of Central (See Figure \ref{fig:composition-rules}). 
With the development of deep learning, recent researchers learn image cropping in a data-driven manner and many aesthetic datasets are constructed to encode the aesthetic preference of humans. 
Recent methods \cite{saliency-wang2017deep,chen2017learning,wei2018good,rebuttal-lu2019listwise} treat it as an object detection task. They used aesthetic datasets to train an aesthetic evaluation model and applied it to compare candidate crops.      
Due to the power of deep learning, these methods have brought progresses in this field, but the intrinsic mechanism remains unrevealed.

In this paper, we propose an interpretable image cropping model to produce both composition-aware and saliency-aware Aesthetic Score Maps, called ASM-Net. Our approach was first inspired by the Class-Activation-Map (CAM) method \cite{CAM-zhou2016learning}, which uses a class activation map to localize the most discriminative image regions in image classification task.
Similarly, we expect to use an aesthetic score map to localize aesthetically important image regions. The aesthetic score of a region can be obtained via average pooling and the regions with larger aesthetic scores are of higher aesthetic quality.
However, direct application of CAM has been proven ineffective because the aesthetic evaluation task is more complicated than classification and one region cannot be simply represented by a single score.
%Inspired by the composition rules, we realize that identifying the relative position of a key object is crucial in evaluating the composition quality of a crop. 
For example, in Rule of Central (left panel in Figure \ref{fig:composition-rules}), we always place the key object at the center to achieve a symmetric and balanced photo crop. So the aesthetic score of the key object located at the center should be larger than off-center.
So we realize that the aesthetic score of a region should vary with its relative position in different crops.

To achieve this goal, we first design some composition patterns to accommodate composition rules so that our model is capable of capturing different composition preferences in the training data.  
Each composition pattern divides a crop into multiple non-overlapping composition partitions (see Figure \ref{fig:composition-pattern}). 
Given a composition pattern, the aesthetic score of a region depends on the composition partition it belongs to in different crops. We name this property as ``composition-aware". In this case, one region is not represented by a single score, but a set of composition-aware scores. Based on such a composition-aware aesthetic score map, we design composition-aware pooling to  calculate crop-level aesthetic scores efficiently.

Furthermore, we believe that our aesthetic score map should focus more on visually salient objects, so we introduce visual saliency into our model as extra supervision. Previous saliency-based image cropping methods  \cite{saliency-wang2017deep,saliency-lu2019aesthetic} generally assume that the most salient object is the most important content and must be included in the best crop. However, we find that this assumption does not hold in many real-world images. For example, landscape photos could have no salient objects, while a party photo might have multiple salient figures that are equally important. These methods could fail in such images.   

Unlike previous saliency-based methods, our assumption is more flexible and practical: a salient region should be more sensitive to the composition partition it belongs to than other regions. We design a novel saliency-aware loss to realize this assumption. By enforcing salient regions to be more composition-sensitive, our model could learn to place salient objects in the proper positions.
%Our experiments also demonstrate the superiority of our assumption.  
With such a composition-aware and saliency-aware aesthetic score map, we could unveil the intrinsic  mechanism of image cropping by locating the aesthetically important regions. Moreover, our model can be easily applied to a wide range of real-world applications. 
The main contributions of our work are as follows:
\begin{itemize}
\item We propose a novel image cropping model ASM-Net with composition-aware and saliency-aware aesthetic score map, which can encode content preference and composition preference.% to learn and encode composition rules and visual saliency in a unified framework. 

\item Our model is able to unveil the intrinsic mechanism of image cropping, by localizing aesthetically important regions based on our aesthetic score map. 

\item Our model outperforms the state-of-the-art methods on three benchmark datasets and has good generality to real-world applications.
\end{itemize}

\section{Related Work}

\noindent\textbf{Aesthetic Evaluation:}
Recently, some large scale aesthetic datasets, like AVA \cite{murray2012ava} and AADB \cite{AesRankNet-kong2016photo}, have been constructed, which enables learning aesthetic evaluation models in a data-driven manner. 
%AVA \cite{murray2012ava} is a large aesthetic dataset that contains rich image-level aesthetic annotations. The AADB dataset \cite{AesRankNet-kong2016photo} provides not only aesthetic scores but also informative attributes. 
By using these datasets, many deep learning methods \cite{aeseval-lu2014rapid,aeseval-lu2015deep,aeseval-hosu2019effective}  have been proposed for image-level aesthetic evaluation.
However, some researchers realized that we need more fine-grained aesthetic annotations for image cropping so that many similar crops could be compared. Thus, \citeauthor{chen2017quantitative} \shortcite{chen2017quantitative} first proposed a dataset to compare randomly sampled crops from one image. 
Following this idea, \citeauthor{wei2018good} \shortcite{wei2018good} increased the number of crops in each image and proposed the first densely annotated image cropping dataset, in which each image has 24 annotated crops with aesthetic scores. \citeauthor{zeng2019reliable} \shortcite{zeng2019reliable} took a further step and presented another densely annotated dataset with about 85 crops for each image.
In our method, we use such datasets to train our image cropping model.

\noindent\textbf{Aesthetic-based Image Cropping:}
Aesthetic-based image cropping improves the cropping results by increasing their aesthetic quality. Early methods \cite{zhang2013weakly,yan2013learning,fang2014automatic} achieve this goal by using hand-crafted features and composition rules. %\citeauthor{fang2014automatic} \shortcite{fang2014automatic} proposed to model the quality of crops with well-composed images. 
By introducing deep learning into this area, some methods proposed to solve image cropping in a data-driven manner.
%\citeauthor{chen2017learning} \shortcite{chen2017learning} proposed to learn from professional photographs paired with its low-quality crop.
\citeauthor{wei2018good} \shortcite{wei2018good} used an object detection framework \cite{liu2016ssd} with two networks, one for generating candidate crops and one for aesthetic evaluation.
More recently, \cite{zeng2019reliable} proposed new metrics to evaluate the performance of image cropping models.
Compared with these deep learning-based approaches, our method is more interpretable. 

\noindent\textbf{Saliency-based Image Cropping:}
The saliency-based methods \cite{saliency-sun2013scale,chen2016automatic,saliency-wang2017deep,saliency-lu2019aesthetic,rebuttal-lu2019end,rebuttal-li2019image} focused on preserving the most important content in the best crop. \citeauthor{saliency-wang2017deep} \shortcite{saliency-wang2017deep} used a saliency dataset \cite{jiang2015salicon} to train a network to generate candidate crops that cover the most salient region. Similarly, \citeauthor{saliency-lu2019aesthetic} \shortcite{saliency-lu2019aesthetic} proposed to generate an initial visual saliency rectangle to include the most important objects.
%, followed by using a regression network to find the best crop.
Unlike these methods, our method has a different assumption on the relationship between visual saliency and aesthetic evaluation. 

\section{Methodology}
\begin{figure*}[htb]
    \centering
    \includegraphics[width=15cm]{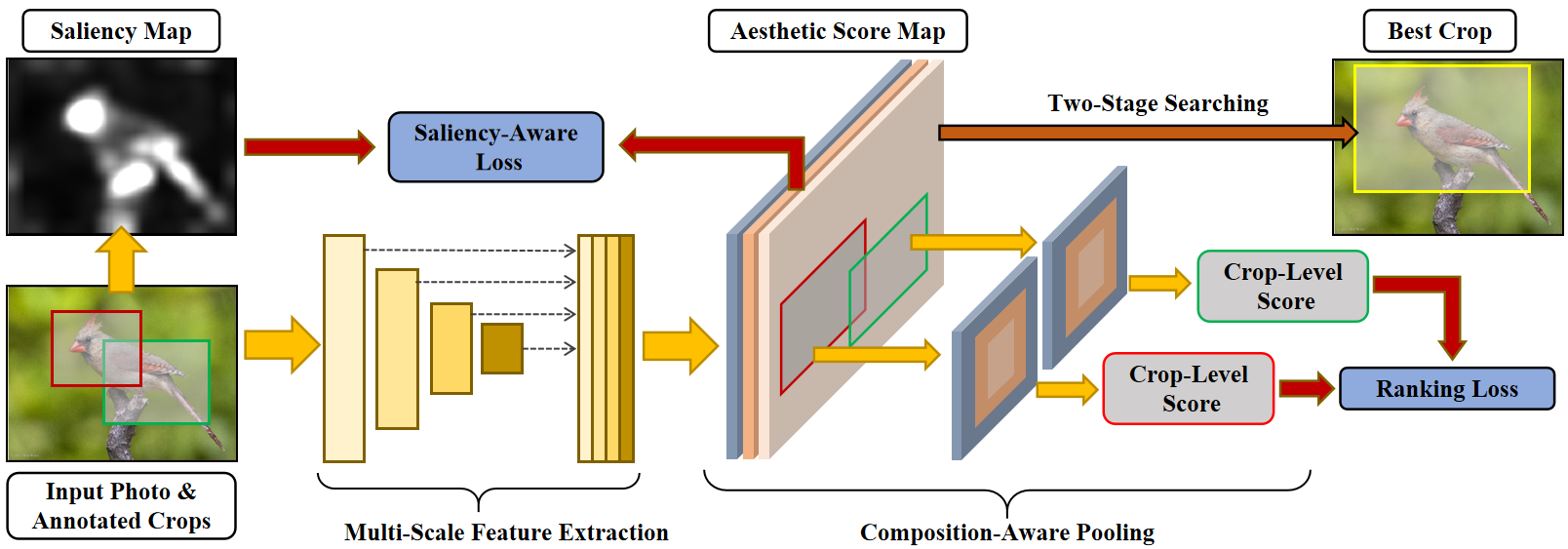}
\caption{Illustration of our model. \textbf{Train stage}: We first obtain the aesthetic score map based on multi-scale features and then produce crop-level scores with composition-aware pooling. Finally, we train our model with  ranking loss and saliency-aware loss.
\textbf{Inference stage}: For each test image, we use the learned model to obtain its aesthetic score map. Then we apply two-stage searching to find the best image crop.}
    \label{fig:model}
\end{figure*}

The flowchart of our method is illustrated in Figure \ref{fig:model}. 
%Our aesthetic score map is both composition-aware and saliency-aware. We train our model with traditional ranking loss and our proposed saliency-aware loss. 

\subsection{Composition-Aware Aesthetic Score Map}
\noindent\textbf{Aesthetic Score Map:} Similar to Class-Activation-Map (CAM)  \cite{CAM-zhou2016learning}, which can localize discriminative regions based on class activation map, we attempt to obtain an aesthetic score map for ease of localizing aesthetically important regions, which can be achieved by Fully Convolutional Network \cite{fully-chen2014semantic,long2015fully}. Considering that a high-resolution score map could be beneficial for region localization and imaging cropping, we upsample the feature maps from different layers to the input image size and concatenate them as our feature map. Then, we apply a $1\times 1$ convolutional layer to transform this feature map to an aesthetic score map, with a larger score indicating higher aesthetic quality.

We use $\mathcal{I} = \{p_{i,j}|0<i\leq H ,0<j\leq W\}$ to denote an $H\times W$ input image with $p_{i,j}$ being its pixel. Besides, we use $\mathbf{M}$ to represent the aesthetic score map of $\mathcal{I}$.
At first, we assume that $\mathbf{M}$ is a $H\times W$ matrix with $m_{i,j}$ being its element. Given an image crop $\mathcal{X}_k \subseteq\mathcal{I}$, we can obtain the aesthetic score of this crop via average pooling $\frac{1}{|\mathcal{X}_k|}
\sum_{p_{i,j}\in\mathcal{X}_k}  m_{i,j}$. Similarly, the aesthetic score of a region in this crop can also be calculated via average pooling. In this case,  one region will get the same aesthetic score in different crops containing this region. However, as discussed in Section~\ref{sec:introduction}, we find that this design is not compatible with aesthetic evaluation. When evaluating the aesthetic quality of a region, we should consider its relative position in different crops. 

\begin{figure}[tb]
    \centering
    \includegraphics[width=6cm]{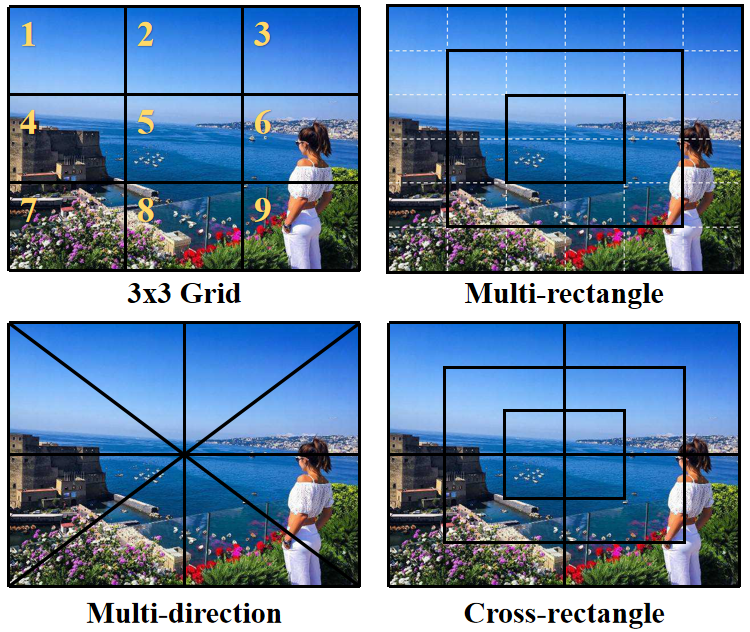}
\caption{Four different composition patterns. \textbf{3x3 Grid}: 9 partitions of the same size. \textbf{Multi-rectangle}: 3 partitions, \textit{center}, \textit{middle}, and \textit{outer}. As the white dotted lines show, it can be divided into a $6\times 6$ grid in implementation. \textbf{Multi-direction}: 8 partitions for 8 directions. \textbf{Cross-rectangle}: 12 partitions, inspired by the camera auxiliary lines.}
\label{fig:composition-pattern}
\end{figure}

Therefore, we propose composition patterns to achieve this goal. Several instantiations of composition patterns are listed in Figure~\ref{fig:composition-pattern}, which share similar spirit with the auxiliary lines used in human-defined composition rules (see Figure~\ref{fig:composition-rules}). We assume that a composition pattern has $L$ non-overlapping composition partitions indexed by $1,2,...,L$.
Then, given a crop and the composition pattern, we can divide this crop into $L$ corresponding composition partitions based on the composition pattern. For each region in this crop, we can easily get its partition index (\emph{i.e.}, $1,2,...,L$) based on its relative position in this crop. The aesthetic score of a region should be aware of its composition partition index in different crops, and we refer to this property as ``composition-aware".

Inspired by R-FCN~\cite{rfcn-dai2016r}, we expand the aesthetic score map to a tensor $\mathbf{M}\in\mathcal{R}^{H\times W \times L}$ with one channel corresponding to one composition partition.
Given an image crop $\mathcal{X}_k \subseteq\mathcal{I}$ and a pixel $p_{i,j}\in\mathcal{X}_k$, we use $\gamma_k(i,j)\in\{1,2,...,L\}$ to denote the composition partition index of $p_{i,j}$ in $\mathcal{X}_k$. So the composition-aware score of $p_{i,j}$ in $\mathcal{X}_k$ is  $\phi_{k}(i,j)=m_{i,j,\gamma_k(i,j)}$. 

\begin{figure}[tb]
    \centering
    \includegraphics[width=8cm]{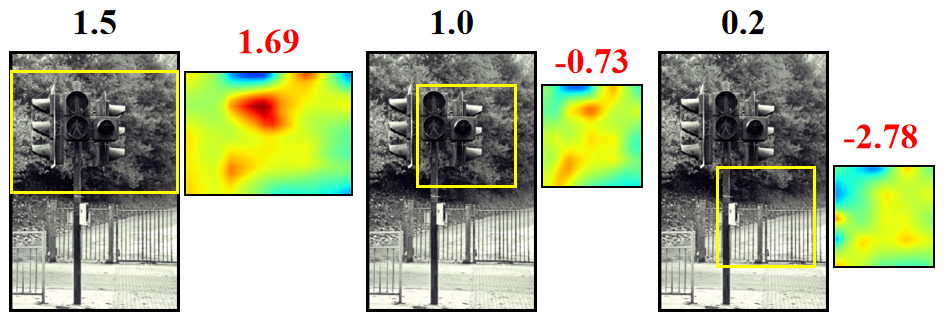}
    \caption{The crop-level aesthetic score maps for three crops in the same image. The annotated scores are in black and the predictions are in red. }
    \label{fig:cs-score}
\end{figure}

\noindent\textbf{Composition-Aware Pooling:}
To calculate crop-level aesthetic scores, we design a composition-aware pooling operation to realize this function.  
We use $\Phi(\cdot)$ to denote composition-aware pooling and thus $\Phi(\mathcal{X}_k)$ is the crop-level aesthetic score of $\mathcal{X}_k$. 
In particular, the crop-level aesthetic score $\Phi(\mathcal{X}_k)$ is defined as the average of the composition-aware aesthetic scores of all pixels in the crop $\mathcal{X}_k$ (see the visualization results in Figure \ref{fig:cs-score}):
\begin{equation}
\Phi(\mathcal{X}_k) \equiv
\frac{1}{|\mathcal{X}_k|}
\sum_{p_{i,j}\in\mathcal{X}_k}  \phi_k(i,j)
\label{equ:crop-score}
\end{equation}
In our implementation, we calculate (\ref{equ:crop-score}) in an efficient way. Given a $h\times w$ crop $\mathcal{X}_k$, we first obtain its crop-level  score map $\mathbf{M}_k\in\mathcal{R}^{h\times w \times L}$ from $\mathbf{M}$. 

Suppose our used composition pattern is $x\times y$  grid pattern, we perform spatial average pooling within each cell of $\mathbf{M}_k$ to get the pooled score map $\mathbf{M}_k^{'} \in \mathcal{R}^{x\times y \times L}$. Note that $L = x\cdot y$ for $x\times y$  grid pattern, so $\mathbf{M}_k^{'}$ can be further reshaped into a square matrix $\mathbf{M}_k^{''} \in \mathcal{R}^{L \times L}$. Then the crop-level score of $\mathcal{X}_k$ is: 
\begin{equation}
\Phi(\mathcal{X}_k) = \frac{1}{L}\sum \text{diag}(\mathbf{M}_k^{''}),
\label{equ:crop-pool}
\end{equation}
in which $\text{diag}(\cdot)$ denotes the diagonal elements.

When dealing with a non-grid pattern with $L$ partitions, we first transform this pattern to a grid pattern with $\tilde{L}$ partitions. In particular, we divide each partition into small cells and each small cell is a new partition (see the multi-rectangle in Figure \ref{fig:composition-pattern} with $\tilde{L}=36$). The multi-direction pattern is a little special but can also be handled in the same way. To be exact, we divide the whole pattern into $12\times 12$ cells ($\tilde{L}=144$) so that each partition is approximately covered by 18 cells.
Then we use $\mathbf{M}\in\mathcal{R}^{H\times W \times L}$ to build a new map $\widetilde{\mathbf{M}}\in\mathcal{R}^{H\times W \times \tilde{L}}$. If the $\tilde{l}$-th new partition comes from the $l$-th old partition, we have $m_{i,j,\tilde{l}} = m_{i,j,l},~\forall~i,j$. 
Then we directly use $\widetilde{\mathbf{M}}$  to replace $\mathbf{M}$ for composition-aware pooling.

\noindent\textbf{Training with Ranking Loss:}
We use $\{(\mathcal{X}_k, y_k)|\mathcal{X}_k\in \mathcal{I},~k=1,...,K\}$ to denote the $K$ annotated crops from $\mathcal{I}$, in which $y_k$ is the annotated score of $\mathcal{X}_k$.
During training, we need to learn $\Phi(\mathcal{X}_k)$ with supervision from $y_k$. Following \cite{chen2017learning,wei2018good}, we learn the relative ranking order of all the crops with the ranking loss, because the ability to correctly ranking crops with similar contents is essential in image cropping \cite{chen2017quantitative}.

For each training image, we first use its annotated crops to generate training pairs based on the differences of the annotated scores.
Specifically, given a threshold $\delta$, the crop pairs are denoted as
\begin{equation}
\mathcal{R}=
\{ (k,t)|y_k-y_t\ge\delta;k\ne t;\mathcal{X}_k,\mathcal{X}_t\in\mathcal{I}\}.
\end{equation}
Since we have totally $K$ annotated crops, we can generate at most $(K^2-K)/2$ crop pairs with $\delta = 0$. 
By increasing $\delta$, we can filter out confusing pairs that might be harmful to training. Then the remaining crop pairs are used to train our model with ranking loss:
\begin{equation}
\mathcal{L}_\text{rank}=
\sum_{(k,t)\in\mathcal{R}}
\max\{0,1+\Phi(\mathcal{X}_t)-\Phi(\mathcal{X}_k)\}.
\end{equation}
%For each crop pair, our model is supposed to predict which one has a higher score correctly. Since the aesthetic score map is shared by all the crops, our model has to utilize the composition partitions to fit the training data. So $\mathcal{L}_\text{rank}$ can make the aesthetic score map become composition-aware.     

\subsection{Saliency-Aware Aesthetic Score Map}
\label{subset:Saliency-Aware}
Before introducing visual saliency to our model, we first 
describe the concept of aesthetically important regions, which is helpful in interpreting image cropping.

\noindent\textbf{Aesthetically Important Region:}
We use the term ``aesthetically important'' to denote the regions that are very influential in image cropping.   
We summarize two types of aesthetically important regions:
1) A region has important content which needs to be preserved, and thus a good crop is supposed to include this region. These ``content-important'' regions will seriously influence the location of good crops.
2) A region is sensitive to its composition partition index. When including such ``composition-sensitive'' region in a crop, they should be carefully placed in proper positions; otherwise they may lead to aesthetic catastrophe. 

In fact, content-important regions and composition-sensitive regions can be explained based on our aesthetic score map. We define two statistics $\theta_\text{avg}(\cdot)$ and $\theta_\text{std}(\cdot)$, in which
$\theta_\text{avg}(p_{i,j})$ is the average aesthetic score of $p_{i,j}$ over all composition partitions and $\theta_\text{std}(p_{i,j})$ is the corresponding standard deviation:
\begin{eqnarray}
\theta_\text{avg}(p_{i,j})&=& \frac{1}{L} \sum_{l=1}^L m_{i,j,l},\\
\theta_\text{std}(p_{i,j}) &=& 
\left(
\frac{1}{L}
\sum_{l=1}^L
(m_{i,j,l}-\theta_\text{avg}(p_{i,j}))^2
\right)^{1/2}.
\end{eqnarray}

Intuitively, a region with high $\theta_\text{avg}$ has a large aesthetic score on average, and good crops are most likely to contain this region. So the regions with high $\theta_\text{avg}$ could be deemed as content-important regions. A region with 
$\theta_\text{std}$ is more sensitive to its composition partition index, because $\theta_\text{std}(\cdot)$ describes the average score difference caused by the change of its belonging composition partition. So the regions with high $\theta_\text{std}$ could be treated as composition-sensitive regions. 
After obtaining an aesthetic score map, we can draw heat maps based on $\theta_\text{avg}$ and $\theta_\text{std}$ to localize aesthetically important regions.

\noindent\textbf{Saliency-Aware Loss:}
\label{subsec:saliency}
Visual saliency is a perceptual quality that can make objects attract more human attention \cite{hou2009dynamic}. It is studied in different applications like saliency prediction \cite{saliency-prediction-pan2016shallow}, salient object detection \cite{borji2015salient-detection}, and video object segmentation \cite{wang2017saliency-video}. In the image cropping task, we can detect the salient regions to preserve the important contents in the best crop. 
Some previous methods \cite{saliency-wang2017deep,saliency-lu2019aesthetic} have already tried in this direction with a simple assumption: the most salient region should be a content-important region. 
On the premise of this assumption, their methods generate candidate crops covering the detected salient region, so that the predicted best crop will not miss the salient region. 
If our model follows this assumption, we should add a constraint that the salient regions have large $\theta_\text{avg}$ during training. 

However, we realize that such an assumption has several limitations. 
First, many real-world images like a party photo could have multiple salient regions, and we cannot  only focus on the most salient one. Second, a landscape photo may not have any salient object, which could result in failure cases as reported in \cite{saliency-lu2019aesthetic}. Third, a salient object could also be a  distraction that should be excluded from a good crop.
%This assumption is also so rigid that the results of image cropping will be strictly limited by the accuracy of saliency detection. 

In our method, we adopt a more reasonable assumption: salient regions should be composition-sensitive, which can be explained as follows.
According to the definition of saliency, a salient region must be visually attractive. When looking at an image, salient regions will draw more attention than others, so people will care more about whether they are placed in proper positions.   
%Using such a common observation, we believe a salient region should be composition-sensitive.  
We first use the saliency detection model \cite{hou2007saliency} to obtain a saliency map $\mathbf{S}$ for image $\mathcal{I}$, in which $s_{i,j} \in [0,1]$ is the saliency score for $p_{i,j}$. Then, we propose our saliency-aware loss $\mathcal{L}_\text{sal}$ by weighing $\theta_\text{std}(p_{i,j})$ with saliency score:
\begin{equation}
\mathcal{L}_\text{sal} =
\frac{1}{HW}
\sum_{p_{i,j}\in \mathcal{I}} 
(1-s_{i,j})
\frac{\theta_\text{std}(p_{i,j})}{\bar{\theta}_\text{std}},
\end{equation}
in which $\bar{\theta}_\text{std}$ is the average of all $\theta_\text{std}(p_{i,j})$ in $\mathcal{I}$.
For non-salient regions, $(1-s_{i,j})$ is large, so they will be penalized for having large $\theta_\text{std}$. For salient regions, $(1-s_{i,j})$ is small, so they are allowed to have large $\theta_\text{std}$. 
Notice that $\theta_\text{std}(\cdot)$ reflects our model capacity. In extreme case, if $\theta_\text{std}(p_{i,j})=0$ for all pixels, our aesthetic score tensor will reduce to aesthetic score matrix with only one channel, and the shrinked model capacity will cause the difficulty of minimizing $\mathcal{L}_\text{rank}$.
By penalizing $\theta_\text{std}$ of non-salient regions, we actually encourage salient regions to have larger $\theta_\text{std}$ in order to fit the training data. In this way, salient objects are encouraged to be composition-sensitive, so that our model can learn to place salient objects in proper positions.

%Essentially speaking, $\mathcal{L}_\text{sal}$ is used to regularize our model to build a strong relationship between the crop-level scores and the composition partition of salient regions like a human will do.  So the intrinsic  mechanism of the aesthetic score map can become more similar to human. We also find this loss can bring improvement on model performance. 

In the training stage, we use images with annotated crops and their saliency maps to train our ASM-Net with the total training loss:
\begin{equation}\label{eqn:total_loss}
\mathcal{L} = \mathcal{L}_\text{rank} + \lambda\mathcal{L}_\text{sal},
\end{equation}
in which $\lambda$ is a trade-off parameter.

\subsection{Two-Stage Search for Image Cropping}
\label{subsec:cropping}
During inference, the saliency map is no longer needed. Given an image, we first use the learned model to produce an aesthetic score map. 
Based on the aesthetic score map, we aim to find the best crop by generating a lot of candidate crops and comparing their crop-level scores. Previous methods \cite{chen2017learning,wei2018good,zeng2019reliable} adopt different strategies in this step. Here, we design a simple yet effective searching strategy to reduce the searching space. 
Specifically, our searching method has two stages.
In the first stage, we conduct a coarse-grained search on a set of pre-defined anchor boxes as in \cite{wei2018good}. 
%The crops with high scores in this stage are  included the most important contents. 
In the second stage, we conduct fine-grained searching around the top crops (\eg, top 10) in the first stage, by slightly varying their center position, size, and aspect ratio. 
%Then the best crop will be the one that has the most delicate composition. 
%Finally, with such a two-stage searching strategy, we will obtain the best crop that has successfully captured the most important content with the best holistic composition.

\section{Experiments}
\begin{table}[tb]
    \centering
\begin{tabular}{|c|c|c|}\hline
Pattern & \#Partitions & Accuracy$\uparrow$ \\\hline
$1\times 1$ grid & 1&  59.46 \\ 
$1\times 3$ grid & 3 &  75.64 \\ 
$3\times 1$ grid & 3 &   71.33 \\  \hline
$5\times 5$ grid & 25 &   86.38 \\
multi-rectangle    & 5 &  81.34 \\
multi-direction    & 8 & 83.22\\
cross-rectangle    & 12 & \textbf{87.59}\\\hline
\end{tabular}
 \caption{Accuracies (\%) of our method with different composition patterns on the validation set of CPC dataset.}
 \label{tab:ablation}
\end{table}

\begin{figure}[tb]
    \centering
    \includegraphics[width =6.5cm]{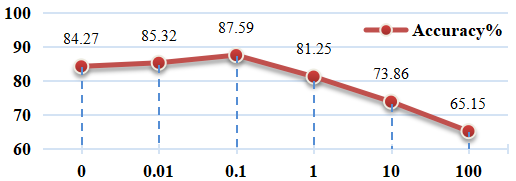}
    \caption{Performance variation of our method with different trade-off parameter $\lambda$ on the validation set of CPC dataset.}
    \label{fig:lambda}
\end{figure}

\label{sec:experiment}
\subsection{Datasets}
We use four image cropping datasets for experiments, two for training and two for testing.

\noindent\textbf{CPC:}
The CPC dataset \cite{wei2018good} is the first densely annotated image cropping dataset. It contains 10797 images, and each image has 24 crops with multiple scores annotated by Amazon Mechanical Turk workers. We use the average annotated score for each crop as ground truth.

\noindent\textbf{GAICD:}
The GAICD dataset \cite{zeng2019reliable} is also a densely annotated dataset, but each image has more annotated crops than the CPC dataset.
It has 1236 images and about 86 crops for each image.

\noindent\textbf{FCDB \& FLMS:}
The FCDB \cite{chen2017quantitative} and FLMS \cite{fang2014automatic} datasets only have annotations for the best crop in each image, which are given by ten annotators. FCDB contains 358 images, and FLMS contains 500 images. These two datasets are only used as the test set in the image cropping task. 
 
\subsection{Implementation Details}

Following the setting in \cite{wei2018good,zeng2019reliable}, we use VGG16 as the backbone model for feature extraction. For generating crop pairs, we have $\delta=0.3$ for CPC dataset and  $\delta=0.9$ for GAICD dataset. 
We choose an unsupervised method in \cite{hou2007saliency} to produce saliency maps for training images. 
%It can highlight all salient regions (not objects) by using Fourier transform and statistical features, which is more suitable for real-world photos that could have multiple salient objects or no salient object.

\subsection{Ablation Study}
We first conduct experiments to study the effectiveness of different composition patterns and our saliency-aware loss. Here, we train our model with the CPC dataset and leave 2000 images as the validation set. We report the ranking accuracy, \emph{i.e.}, the accuracy of correctly ranking the generated crop pairs, on the validation set of CPC dataset. 

\noindent\textbf{Composition Pattern:}
We explore four types of composition patterns in Figure \ref{fig:composition-pattern} and the results are reported in Table \ref{tab:ablation}.
%Due to the limited space, we only report 3 representitive results and 4 best results for each type in Table \ref{tab:ablation}.
First, we investigate the grid pattern with different numbers of partitions.
When using $1\times 1$ grid, our model is no longer composition-aware because it will assign a region with the same score regardless of its different composition partition indices in different crops. 
When using  $1\times 3$ (\emph{resp.}, $3\times 1$) grid, our model is only aware of horizontal (\emph{resp.}, vertical) shift. We observe that $1\times 3$ gird has higher accuracy, which means that horizontal shift has more impact on the aesthetic quality of a crop, which has also been proved by empirical studies \cite{palmer2013visual,abeln2016preference}.
Then, we compare four types of composition patterns and report the best results for four types of composition patterns. 
We conjecture that their performance differences are attributed to their sensitivity to different types of position shifts. To be exact, multi-rectangle and multi-direction are only sensitive to one type of position shift (\ie, in-out shift or directional shift), which restrains them from capturing more complicated composition rules in the training data, so they have lower results. The best result comes from the cross-rectangle, which is like a combination of multi-rectangle and multi-direction. It is inspired by camera auxiliary lines, which proves to be more compatible with learning composition rules. Due to the competitive performance of the cross-rectangle, we use it as the default composition pattern for the rest of the experiments.

\noindent\textbf{Saliency-Aware Loss:}
Recall that we have a trade-off parameter $\lambda$ before our saliency-aware loss in (\ref{eqn:total_loss}). We vary $\lambda$ in the range of [0, 100] and report the results in Figure \ref{fig:lambda}. When we do not use the saliency-aware loss  ($\lambda$=0),  the accuracy drop is 3.22\% compared with the best accuracy achieved when $\lambda$=0.1. When $\lambda$ becomes larger, the accuracy begins to drop because it will reduce our model capacity, as discussed in Section~\ref{subset:Saliency-Aware}. We use $\lambda$=0.1 by default for the rest of experiments.

\begin{table}[tb]
    \centering
    \begin{tabular}{|l|l|l|}\hline
\multicolumn{1}{|c|}{Method} &  \multicolumn{1}{|c|}{IoU$\uparrow$}& \multicolumn{1}{|c|}{Disp$\downarrow$}  \\ \hline
RankSVM \cite{chen2017quantitative}  &  0.6020 &    0.106\\
AesRankNet \cite{AesRankNet-kong2016photo}    & 0.4843 &    0.140 \\
MNA-CNN    \cite{MNA-CNN-mai2016composition} & 0.5042    & 0.136\\
VFN    \cite{chen2017learning} &0.6842 &    0.084\\
VPN    \cite{wei2018good} & 0.7109 &    0.073\\
VEN \cite{wei2018good}    & 0.7349  &    0.072 \\ \hline
ASM-Net (ours) &\textbf{0.7489}& \textbf{0.068}\\ \hline
\end{tabular}
    \caption{Results of different methods on the FCDB dataset.}
    \label{tab:FCDB}
\end{table}

\begin{table}[tb]
    \centering
    \begin{tabular}{|l|l|l|}\hline 
\multicolumn{1}{|c|}{Method} &  \multicolumn{1}{|c|}{IoU$\uparrow$}& \multicolumn{1}{|c|}{Disp$\downarrow$}  \\ \hline
\citeauthor{chen2016automatic} \shortcite{chen2016automatic}  &    0.6400&    0.075 \\
\citeauthor{fang2014automatic} \shortcite{fang2014automatic} &0.7400&     - \\
\citeauthor{suh2003automatic} \shortcite{suh2003automatic} &0.7200&    0.063 \\
ABP+AA \cite{saliency-wang2017deep}&    0.8100&    0.057 \\
VPN    \cite{wei2018good} &0.8352&    0.044  \\
VEN    \cite{wei2018good} &0.8365&    0.041  \\ \hline
ASM-Net (ours) & \textbf{0.8486}& \textbf{0.039} \\ \hline
\end{tabular}
    \caption{Results of different methods on the FLMS dataset.}
    \label{tab:FLMS}
\end{table}
\subsection{Performance on Image Cropping}
Image cropping models are usually evaluated based on single best crop prediction. However, \citeauthor{zeng2019reliable} argue that image cropping is naturally a subjective and flexible task without a unique
solution, and thus suggest evaluation based on multiple crop ranking. 
Here we conduct experiments based on both best crop prediction and multiple crop ranking.

\noindent\textbf{Predicting Best Crop:}
In this task, we predict the best crop and compare it with the ground truth crop.
We follow the experiment setting in  \cite{chen2017learning,wei2018good} and use two metrics for performance evaluation: intersection-over-union (IoU) and boundary displacement (Disp).
We use the CPC dataset as the training set, and FCDB and FLMS datasets as two test sets. The results of our ASM-Net and baseline methods are reported in Table \ref{tab:FCDB}\&\ref{tab:FLMS}.
On FLMS, we beat the saliency-based approach \cite{saliency-wang2017deep}. 
Our method achieves the best performance on both FCDB and FLMS datasets.

\begin{table}[tb]
    \centering
    \begin{tabular}{|l|l|l|l|}\hline
\multicolumn{1}{|c|}{Method}   & SRCC$\uparrow$ & $\text{Acc}_5\uparrow$ & $\text{Acc}_{10}\uparrow$ \\ \hline
VFN \cite{chen2017learning}& 0.450&   26.7  &    38.7\\ 
VEN \cite{wei2018good} &0.621    &37.6    &50.9\\
GAIC \cite{zeng2019reliable}& 0.735&    50.2 &    68.5
  \\ \hline
ASM-Net (ours) &\textbf{0.766}&  \textbf{54.3}&\textbf{71.5}
  \\ \hline
\end{tabular}
    \caption{Results of different methods on the GAICD dataset.}
    \label{tab:GAICD}
\end{table}

\noindent\textbf{Ranking Multiple Crops:}
The above experiments are designed for locating a single best crop, where only the prediction of the best crop is evaluated. In the densely annotated dataset, each image has multiple annotate crops, so \citeauthor{zeng2019reliable} \shortcite{zeng2019reliable} proposed to  compare the ranking order of all annotated crops with two different metrics: Spearman’s rank-order correlation coefficient (SRCC) and generalized top-N accuracy ($\text{Acc}_N$).
While IoU and Disp only care about the best crop prediction, SRCC and $\text{Acc}_N$ evaluate the performance of correctly ranking multiple annotated crops.
Following \cite{zeng2019reliable}, we choose 1036 images in GAICD dataset for training and the remaining 200 images for testing. Many previous methods can only predict the best crop, so they are not applicable to this task.
The results are summarized in Table \ref{tab:GAICD}, from which we observe that our method beats all the applicable baseline methods. 
%Compared with GAIC who presented this GAICD dataset, we still have a large improvement.

\noindent\textbf{Discussion:}
The main advantage of our method is that the aesthetic score map is both composition-aware and saliency-aware. 
Remember that a good image crop is achieved by placing important content with a good composition, so these two properties are very important in the image cropping task. 

With saliency-aware property, our model can focus on the salient regions, which are more likely to contain important content.
Compared with the saliency-based approach (\cite{saliency-wang2017deep}), our method has a more reasonable assumption that is more compatible with aesthetic evaluation.
With composition-aware property, our model is able to encode complicated composition rules learned from training data. Especially when ranking the overlapping crops which are visually similar, our method gains an edge by referring to the composition partition index of their common content. 
%While some previous methods (\cite{chen2017learning,wei2018good}) use neural network models with a fully connected layer for aesthetic evaluation,  their models are insensitive to position shift of image content, so they are not good at comparing different compositions. 
In summary, our model is more compatible with the image cropping task, so it can make better use of the training data and have better performance on both tasks.

\subsection{Application}
\begin{figure}
    \centering
    \includegraphics[width=8.5cm]{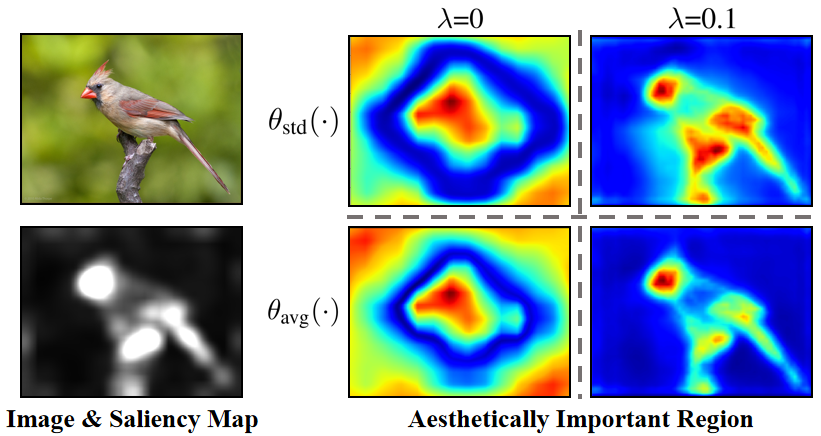}
    \caption{Illustration of aesthetically important regions based on the heat maps of $\theta_\text{std}(\cdot)$ and $\theta_\text{avg}(\cdot)$ with $\lambda$=0 and $\lambda$=0.1.}
    \label{fig:important-regions}
\end{figure}

\noindent\textbf{Localizing Aesthetically Important Region:}
Following the discussion in Section \ref{subset:Saliency-Aware}, we locate aesthetically important regions by drawing heat maps with $\theta_\text{avg}(\cdot)$ and $\theta_\text{std}(\cdot)$ in Figure \ref{fig:important-regions}.
We compare the heat maps between $\lambda$=0 and $\lambda$=0.1.  
We can observe that when not using saliency-aware loss ($\lambda$=0), the image corners are aesthetically important.
We believe it is caused by the spatial bias in the training images.
%We conjure that a model with saliency-aware loss tends to learn more spatial bias in the training data.

\begin{figure}
    \centering
    \includegraphics[width=8cm]{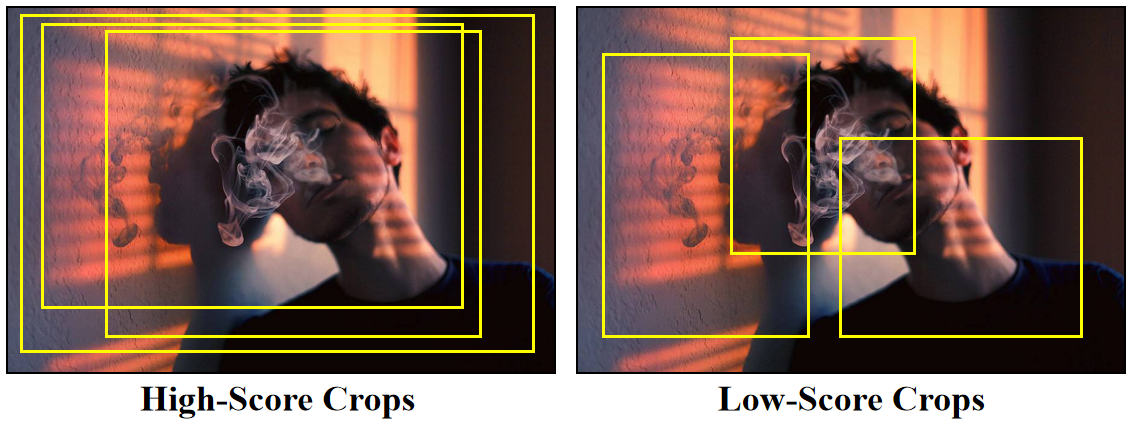}
    \caption{Illustration of spatial bias. High-score crops are prone to cover four corners of the whole image.}
    \label{fig:data-bias}
\end{figure}

The spatial bias means that the annotated score of a crop is related to its position in the whole image. 
For example,  since most photos will place the important contents at the center, the high-score crops are usually central and large enough to cover them, so they are more likely to cover the four corners of the image compared with low-score crops (see Figure \ref{fig:data-bias}). So the model is prone to assign high aesthetic scores to the corners of the image.
In fact, a previous method GAIC has leveraged such statistical prior information to reduce low-score crops in candidate crop generation.
By using saliency-aware loss, we are actually preventing the model from overfitting such spatial bias, so the aesthetically important regions are  closer to the salient regions.

\noindent\textbf{Diverse Image Cropping:}
To demonstrate the ability of our model, we visualize our cropping results in three typical scenes: one salient object, multiple salient objects, and no salient object.   
We first show the best cropping result in three typical scenes (two images for each scene) with five aspect ratios (9:16, 3:4, 1:1, 4:3, and 16:9) in Figure \ref{fig:cropping-results} (see more results in Supplementary).
Notice that a saliency-based method might fail in the scene with no salient object, as reported in \cite{saliency-lu2019aesthetic}.
However, the results have demonstrated that our model can generally produce satisfactory results in different scenes with different aspect ratios.  

\begin{figure}[tb]
    \centering
    \includegraphics[width=8cm]{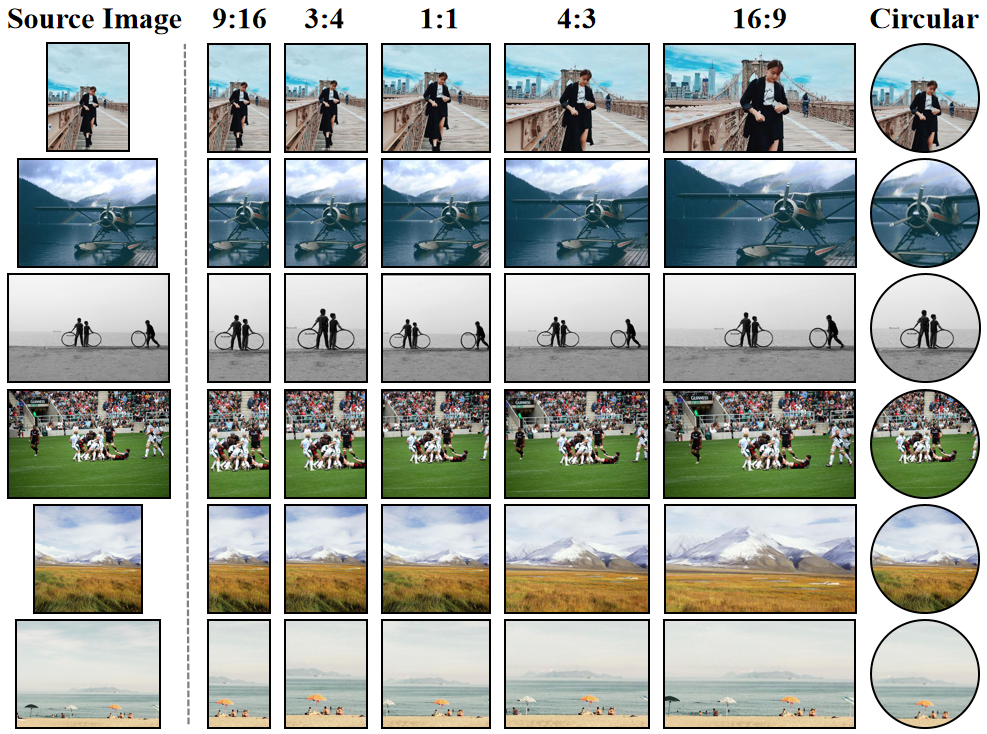}
    \caption{Image cropping results on six images (left column) with different aspect ratios and shapes.}
    \label{fig:cropping-results}
\end{figure}
\noindent\textbf{Circular Cropping:}
Since we obtain crop-level aesthetic score by pooling the aesthetic score map, our model can be naturally used to evaluate image crops of arbitrary shape, by slightly changing the composition-aware pooling operation. 
Now we extend the rectangle cropping task to circular cropping since it can be used in real-world applications like generating circular user icons for websites.
We use $\mathcal{X}_t\subset\mathcal{I}$ to denote a candidate circular crop, which is also the inscribed circle of a square crop $\mathcal{X}_k$. 
During training, we still use rectangle crops due to the lack of annotated circular crops. During prediction, we use $\phi_k(i,j)$ from $\mathcal{X}_k$ as the composition-aware score of $p_{i,j}$ but only pool from the inscribed circle $\mathcal{X}_t$, so the circular crop-level score of $\mathcal{X}_t$ is :
\begin{equation}
\Phi^{'}(\mathcal{X}_t) = 
\frac{1}{|\mathcal{X}_t|}
\sum_{p_{i,j}\in\mathcal{X}_t } \phi_k(i,j).
\label{equ:circle-crop-score}
\end{equation}
With the circular crop-level score defined above, our model finds some best circular crops, which are shown in Figure \ref{fig:cropping-results} (see more results in Supplementary). 
The results demonstrate the generality of our proposed model. 
Since we have not trained the model with any annotated circular image crops, it also proves that our model has mastered a universal aesthetic evaluation criterion that can be shared among crops with different shapes. 

\section{Conclusions}
In this paper, we have proposed an interpretable model to unveil the intrinsic mechanism of the image cropping model.  
Our model can produce an aesthetic score map, in which the aesthetic score of a region is aware of its composition partition index in different crops. 
Moreover, by introducing visual saliency, our model can learn to place salient objects in proper positions.
Our model has achieved the best performance in image cropping tasks and demonstrated good generality in real-world applications.

\section*{Acknowledgement}
The work is supported by the National Key R\&D Program of China (2018AAA0100704) and is partially sponsored by Shanghai Sailing Program (BI0300271) and Startup Fund for Youngman Research at SJTU (WF220403041).

\begin{small}
\bibliographystyle{aaai}
\bibliography{ref}
\end{small}

\end{document}